\documentclass[letterpaper, 10 pt, conference]{ieeeconf}  

\IEEEoverridecommandlockouts                              

\usepackage{graphics} 
\usepackage{epsfig} 
\usepackage{mathptmx} 
\usepackage{times} 
\usepackage{amsmath} 
\usepackage{amssymb}  
\usepackage{graphics} 
\usepackage{graphicx}
\usepackage[ruled]{algorithm2e}
\usepackage{subfigure}
\usepackage{tabu}
\usepackage{booktabs}
\usepackage{multirow}
\usepackage{footnote}
\usepackage{threeparttable}
\usepackage{booktabs}
\usepackage[table]{xcolor}
\usepackage{mwe}
\usepackage{caption}
\usepackage{xcolor}
\usepackage{lipsum}
\usepackage{float}
\usepackage{todonotes}
\usepackage{color,soul}
\usepackage{hyperref}
\usepackage{multirow}
\usepackage{cuted}
\usepackage{stfloats}
\usepackage{amsmath} 
\usepackage{amssymb}  
\usepackage{array}

\title{\LARGE \bf
	Design and Implementation of A Novel Precision Irrigation Robot Based on An Intelligent Path Planning Algorithm}
\author{
	Minghan Chen$^{1}$, Yilong Sun$^{1}$, Xueqing Cai$^{1}$ , Boyi Liu$^{2}$ and Tenglong Ren$^{3}$
	\thanks{This paper was recommended for publication by Editor upon evaluation of the Associate Editor and Reviewers' comments. This work was supported by the National Natural Science Foundation of China, National Natural Science Foundation of Hainan Province and the foundation of Robot and Artificial Intelligence Association for Hainan University awarded to authors.}
	\thanks{$^{1}$Minghan Chen, Yilong Sun, Xueqing Cai are with Hainan University. {\tt\small mh.chen@hainu.edu.cn};
		{\tt\footnotesize yl.sun@hainu.edu.cn};{\tt\footnotesize xq.cai@hainu.edu.cn}}
	\thanks{$^{2}$Boyi liu is with the University of Chinese Academy of Sciences.{\tt\footnotesize by.liu@ieee.org}}
	\thanks{$^{3}$Tenglong Ren is with the Tianjin University. {\tt\footnotesize tenglongren@tju.edu.cn}}
}

\begin{document}
	\maketitle
	\thispagestyle{empty}
	\pagestyle{empty}
    	\begin{strip}
       	\centering
       	\vspace{-2.2cm}
       	\includegraphics[width=1\linewidth]{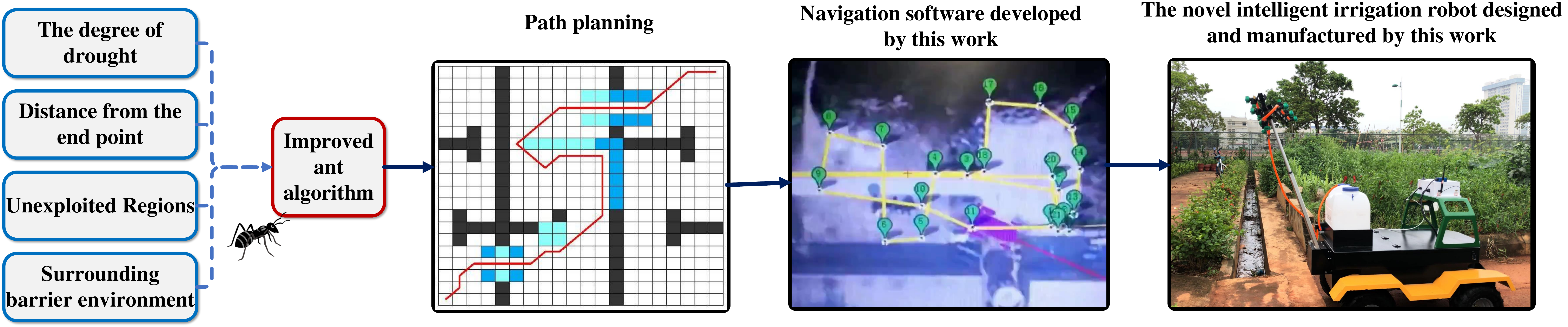}
       	\captionof{figure}{The working structure of the intelligent agricultural robot. We considers the unexplored areas, the distance to the target point, the condition of surrounding obstacles and the degree of regional drought before  path planning. Then, we use the improved algorithm to generate precise irrigation areas and routes. After that, it uploads the results to the navigation system. Finally, the navigation system controls the robot to perform irrigation task.}
       	\label{fig:architecture}
       \end{strip}
	\begin{abstract}
    The agricultural irrigation system is closely related to agricultural production. There are some problems in nowadays agricultural irrigation system, such as poor mobility, imprecision and high price. To address these issues, an intelligent irrigation robot is designed and implemented in this work. The robot achieves precise irrigation by the irrigation path planning algorithm which is improved by Bayesian theory. In the proposed algorithm, we utilize as much information as possible to achieve full coverage irrigation in the complex agricultural environment. Besides, we propose the maximum risk to avoid the problem of lack of inspection in certain areas. Finally, We carried out simulation experiments and field experiments to verify the robot and the algorithm. The experimental results indicate that the robot is capable of fulfilling the requirements of various agricultural irrigation tasks.
	\end{abstract}
	\section{INTRODUCTION}
   Agricultural irrigation system determines the nutritional supply status of crops, which in turn affects crop yields. The existing irrigation system mainly adopts basin irrigation[1], sprinkler[2], Internet of things technology[3] et al. However, these methods are either inadequate for saving water, demand high for the topography, or have poor mobile performance. Nowadays, numerous researches focus on 
   addressing these issues. What's more, large-scale agricultural machines can hardly be applied to various agricultural environment. As a result, small robots that complete a variety of agricultural tasks are significantly needed[4]. In this paper, we improve the water resource utilization and realize the full coverage agriculture cruise by designing and implementing an intelligent robot with diverse functions. 
   
   Autonomous navigation is regarded as the basic function of robots[5]. Likewise, the key of the proposed robot is the irrigation path planning algorithm. At present, many path planning algorithms in robots are set without considering the road complexity, and serious impact of drought. Extra manpower are needed working as supervisors and assistants. So the improved algorithm adopts Bayesian theory to consider as many agricultural factors as possible. For instance, the unexplored region, the distance to the target point, the condition of surrounding obstacles as well as the degree of drought. The robot takes advantage of the information gained in the previous cruises and then generates a posterior probability to adjust parameters. Though constant information acquisition and self-adjustment, the intelligent robot is adapted to different agriculture environments rapidly. To evaluate the validation of this algorithm, we carry out simulated experiments and field tests. The experimental results indicate that the algorithm improves the adaptability and intelligence of the robot. 
              
   In conclusion, the contributions of this paper are as follows:
   	\begin{itemize}
   	\item We design and implement an intelligent irrigation
 robot. It is capable of expanding the working area and reducing the water waste at the same time.
   	\item In the work, we propose an improved algorithm based on Bayesian theory. It enables the robot to irrigate precisely.
   	\item We develop an irrigation navigation software to receive results from the algorithm and control the robot.
      \end{itemize}
\section{RELATED WORK}
The shortage of water resources has already became an obstacle to human development. According to a recent survey, the water demand has exceeded supply in more than 40\% of the area. The use in agricultural irrigation accounts for 75\% of the world's fresh water supply[6]. As a result, how to improve the utilization of water resources and reduce manpower input have become concerns of researchers. Some of them attempted to evaluate irrigation areas through decision support systems to design specific irrigation schemes[7]. However, considering the land condition has been continuously changing, the method was lack of adaptability. Some researchers revolutionized agricultural irrigation systems. For example, Angelopoulos et al. developed an intelligent irrigation system based on a wireless sensor network[8]. They were able to monitor soil conditions constantly by embedding sensors. Yet, this method was difficult to be adapted to large-scale planting areas. What's worse, the embedding of sensors affected the growth of plants. 

In 2016, Kushwaha et al. provided new ideas for the implementation of precision irrigation. They introduced the application of robotics in agriculture[9]. O.P. Bodunde et al. designed an adaptive robot based on ZigBee technology in 2019. It was capable of alerting users to water the crops, which was non-sensors and costs less. However, they were not portable and did not really alter the way of manual irrigation.

Afterwards, the development of mobile irrigation robots[10] changed the situation to some extent. But it can only irrigate specific areas based on a preset amount of water. We found out that the existing research results had the mutual disadvantages of poor adaptability and high dependence on manpower. To make the robot more intelligent, we should improve the performance of path planning algorithm. Zuo et al. proposed a full coverage path planning method for agricultural robots based on sub-regions. This method used the Depth First Search (DFS) to look for the coverage sequence of sub-regions. Then, it selected the longest part to avoid making turns in the narrow area[11]. In 2015, I.A.Hameed et al. estimated the total hop/overlap area to make up for the loss of important information caused by the neglect of ground undulation. This method is capable of find the best driving angle to minimize the hop over overlapping areas so that the economic impact can be reduced[12].Mogens Graf plessen improved the path planning according to the special need of repeatedly returning to the site entrance to fill the seed package when planting the wheat and other crops.
 Different from the above approaches, Mogens et al. developed a path planning algorithm for the specific crop planting[13].To sum up, high dependence on manpower and the difficulty of saving water resource have become widespread problems in irrigation systems. Therefore, continuous improvement is necessarily needed to minimize the human participation and maximize the water utilization in the irrigation process.
 
 There are many paths for the robot to reach the target point from the initial position. Generally, the robot were designed to choose the path that achieved the shortest distance or the shortest time. However, few researches work on path planning in precision irrigation. In this paper, we design and implement an intelligent robot and then propose an algorithm. We take the unexplored region as the main influencing factor. The secondary factors are the distance to the target point and the condition of surrounding obstacles. The degree of regional drought acts as the auxiliary factor. Then, we make path planning decisions to achieve multiple information collection and full coverage cruise of agriculture environment. 

\section{METHODOLOGY}
In this section, we will introduce the irrigation planning algorithm in detail, which includes the concepts and formula in Bayesian theory, the application of maximum risk and the specific steps of operation.

\subsection{Framework}
\begin{figure}[thpb]
	\centering
	\includegraphics[width=1\linewidth]{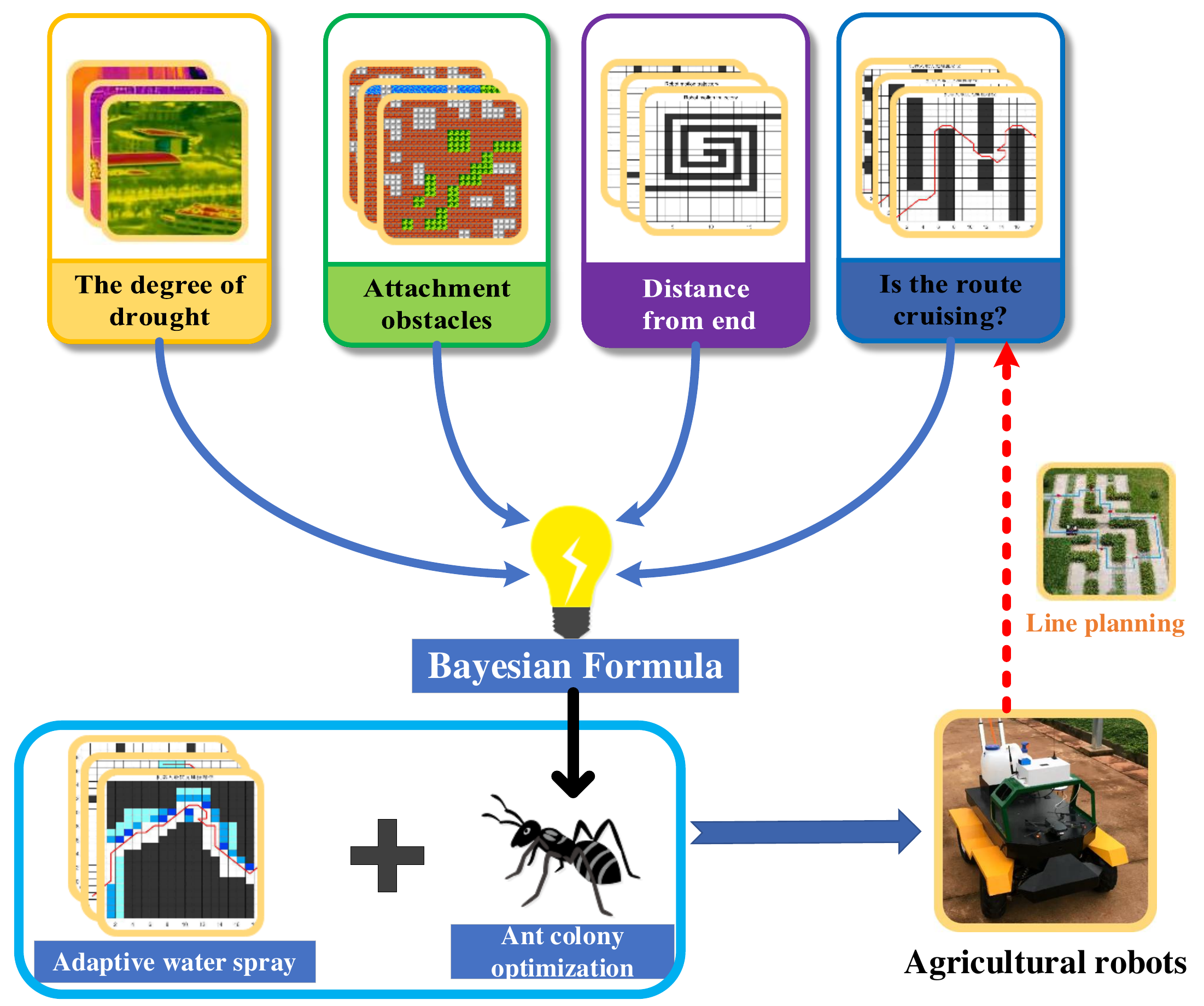}
	\caption{The framework of the improved algorithm. The robot records the degree of drought, the condition of obstacles and the explored areas. Then, it updates the data in the algorithm to plan the next path. Finally, the robot conduct the task according to the instruction.}
	\label{fig:architecture}
\end{figure} 
 The conventional means have achieved good results for trapezoidal obstacles. But they ignored some significant features of precise irrigation. Fig.2 presents the framework of the improved algorithm. Firstly, agricultural robots have to traverse almost every path on the map. Moreover, it is required to accomplish multi-tasks so that it can monitor the whole growing environment of crops accurately. In this work, we consider as many factors as possible and then set parameters for the these environmental factors based on their importance. The unexplored area comes top of the list, followed  by the distance to the target point, the condition of surrounding obstacles and the degree of drought in turn. The robot makes use of the information acquired in the cruise to adjust parameters. However, we discover that certain areas can be dismissed by the robot because the influence of some factors is too small. So We propose the maximum risk to address this problem. It aims at achieving priority irrigation in unexplored areas by adding maximum weight to them after several cruises.

\subsection{Prior probability}
 In the ant colony algorithm, we suppose that $S\text{=}\left\{ {{X}_{1}},{{X}_{2}},{{X}_{3}}\cdots {{X}_{n}} \right\}$ is the set of notes passed by ants. The probability of finding the global optimal solution is marked as $P(S)$. In the feasible region, the sum of the transition probabilities of the nodes is 1. The prior probability is determined by the transition probability. Normally, we choose the note with a higher transition probability, which turns out to be not a perfect choice. Therefore, we introduce the posterior probability for further analysis.

\subsection{Conditional probability } 
The traditional ant colony algorithm only considers the distance between the candidate node and the final target point. When taking multiple factors into consideration, we choose probabilities under certain conditions to represent their impacts on candidate nodes.
$P(S|{{W}_{i}})$ represents the probability of achieving the globally optimal solution after selecting the next node.
\begin{table}[htbp]
	\vspace{6pt}
		\caption{Variables in this section}
		\setlength{\tabcolsep}{1mm}
	\begin{tabular}{m{0.15\columnwidth}m{23.11em}}
    \toprule
		\textbf{variable}&\centering{\textbf{Meaning}} \cr                       
		\hline
		$\overrightarrow{{{X}_{i}}{{X}_{j}}}$       &  The current position of Ant $a$ is ${{X}_{i}}$, the next selected node is ${{X}_{j}}$. $\overrightarrow{{{X}_{i}}{{X}_{j}}}$ is the motion direction of Ant $a$.\\   
		\hline
		$FW({{X}_{j}})$       & Under the grid scale of environment $n\times n$ ,  with candidate ${{X}_{j}}$ as the end point, the current position  ${{X}_{i}}$ as the starting point, the direction of vector $\overrightarrow{{{X}_{i}}{{X}_{j}}}$ is diagonal direction, the square with a side length of $\min \left( \left\lceil \sqrt{n} \right\rceil +1,5 \right)$is the prediction area ${{X}_{j}}$ of the candidate node $FW({{X}_{j}})$, whether the area explored is the main factor, the distance from the target point, the surrounding obstacle environment as a secondary factor, The degree of regional pests and diseases is a secondary factor.  \\ 
		\hline		$Zdx({{X}_{j}})$       & In the prediction area $FW({{X}_{j}})$, the total number of consecutive obstacles in the node ${{X}_{j}}$ range is recorded as $Znum({{X}_{j}})$, $Zdx({{X}_{j}})$ is the total size of the continuous obstacle, and $Zdx({{X}_{j}})=\sum{L\left( Z \right)}$.                                                                                                                                                                    \\\hline
		$Wdx({{X}_{j}})$       & The total size of the untraversed area in the prediction range $FW({{X}_{j}})$ of node ${{X}_{j}}$ is recorded as $Wdx({{X}_{j}})$.                                                                \\ \hline   
		$GHdx({{X}_{j}})$       &The total size of the drought area in the prediction range $FW({{X}_{j}})$ of the node ${{X}_{j}}$ is recorded as $GHdx({{X}_{j}})$, and the initial value of $GHdx({{X}_{j}})$ is 0
		\\ \toprule    
	\end{tabular}

	\label{bs2}
\end{table}

\subsection{Maximum Risk and Posterior Probability} 

There are four influencing factors in the algorithm. Some paths, although they are featured by drought, fail to be precisely irrigated because of the large total weight of the remaining three factors. Therefore, we propose the maximum risk, which is inspired by the minimum risk in Bayesian theory.\\
(1) Application of maximum risk:  we default the degree of drought in the unexplored path to the maximum value. As a result, the robot will give top priority to irrigate this area in the next round. Further more, decisions related to this state are assigned a smallest risk value in the table.
Determine the value of intermediate elements:
$$P({{W}_{i}}|S)=\frac{P\left( {{W}_{i}} \right)P\left( S|{{W}_{i}} \right)}{\sum\limits_{i=1}^{n}{P\left( {{W}_{i}} \right)P\left( S|{{W}_{i}} \right)}}$$
$P({{W}_{i}}|S)$ obtains the posterior knowledge. The algorithm determines the candidate node by posterior probability.\\
(2) calculate the condition risk for each decision \, ${{\alpha }_{i}}, i=1\cdots b$.  
\[R\left( {{\alpha }_{i}}/x \right)=\sum\limits_{j=1}^{c}{{{\lambda }_{ij}}P({{w}_{j}}/x),i=1,\cdots b}\]\\
(3) Select the decision with the minimal conditional risk and determine the object to be identified
\[R\left( {{a}_{k}}/x \right)=\underset{i=1\cdots b}{\mathop{\min }}\,R({{a}_{i}}/x)\]
When the non-traversed path is not included in the optimal solution, we will update the decision table with a smaller value, and repeat the process of calculation and decision-making.
(4)The specific operations are the following:

Step 1: The robot carry out precise irrigation for the first time. The judgment standard includes only the distance to the end point and the condition of surrounding obstacles. The target point is determined according to the posterior probability expression.

Step 2: Divide a circle into cubes, whose diameter is the distance between the two points.Then, we build a data table for all cubes. If it has irrigated before, we will reduce its value by 1. Meanwhile, anther data table is built to record the degree of drought. If the sensor installed identifies that it reaches a certain degree, we will add 1 to the value. The sum of the corresponding values of these two data tables are marked by $Wdx({{X}_{j}})$ and $GHdx({{X}_{j}})$respectively

Step 3: The second round of cruise starts after the first time, when the remaining two factors are added into decision-making. In order to choose the non-cruised route preferentially, we use the maximum risk to increase its probability.

As shown in Algorithm 1, it is the implementation of program. With considering more features above, the proposed algorithm is more applicable for agricultural environment, and can meet the requirements of precision spraying, automatic cruise. The next step is to determine the parameters for better adaptation.

 \begin{algorithm}
		\caption{the improved ant colony algorithm based on Bayesian Theory}
         The set of notes in the generated path S=${{x}_{1}},{{x}_{2}},\cdots, {{x}_{n}}$.
         
		\KwIn {$n$:the round of cruise; ${\lambda }_{i}(i\in 1,2,3,4)$:weights of environmental factors.}
		\While{$ m\ne n$}
		{
			Obtain the value of $GHdx({{w}_{i}}) $ and $ Wdx({{w}_{i}}) $ in  $FW({{w}_{i}})$\\
			${{f}_{1}}({{w}_{i}})=\frac{1}{dist({{w}_{i}},Object)}$\\
			\If{ $Znum({{w}_{i}})=0$}
			{
			{${{f}_{2}}({{w}_{i}})=1$}
			}
        	\If{$Znum({{w}_{i}})\ne0$}
			{
			${{f}_{2}}({{w}_{i}})=\frac{1}{Znum({{w}_{i}})+Zdx({{w}_{i}})}$
			}
		    ${{f}_{3}}({{w}_{i}})=\frac{GHdx({{w}_{i}})}{mS({{w}_{i}})}$\\
            ${{f}_{4}}({{w}_{i}})=\frac{Wdx}{mS({{w}_{i}})}$\\
            $P'\left( S|{{W}_{i}} \right)={{\lambda }_{1}}{{f}_{1}}({{W}_{i}})+{{\lambda }_{2}}{{f}_{2}}({{W}_{i}})+{{\lambda }_{3}}{{f}_{3}}({{W}_{i}})+{{\lambda }_{4}}{{f}_{4}}({{W}_{i}})$\\
            $P(S|{{w}_{i}})=\max \{p'(S|{{w}_{i}})\}$ \\
            set ${{w}_{i}}$ as the starting point, then reselect the candidate note.\\
            $m=m+1$
		}
	\end{algorithm}
\section{EXPERIMENTAL RESULTS AND ANALYSIS}
 In this section, we introduces the experimental setup and results. To verify the effectiveness of the improved algorithm the robot, we need to answer three questions: 1) Can the improved algorithm be applied to diverse farmland environments? 2) How to set the parameters in the model and how they affect the results. 3) Can this algorithm meet the intelligence of the robot for agricultural activities? Therefore, we conducted a simulation experiment and a field test in the agricultural park with a self-designed robot.
\subsection{Simulation experiment}
 In this work, we carry out agricultural simulation experiments to verify the feasibility and intelligence of the algorithm. The simulation experiment requires that the real map can be transformed into the grid map needed for path planning. We conducted experiments with $Matlab$. During the first round of path planning, the relevant factors affecting path planning are the condition of obstacles and distance to the target point. We set a square with a side length of $ min[\sqrt{n}+1.5]$ as the prediction area. The coefficients of ${{f}_{1}}({{w}_{i}})$ and ${{f}_{2}}({{w}_{i}})$  are initialized to ${{\lambda }_{1}}\text{=}{{\lambda }_{2}}\text{=}0.5$. The setting of the coefficient is only related to whether priority is to avoid more obstacles or avoid large obstacles.
 
 As Fig.3 presents, we will use gray, red, green and blue squares representing obstacles, roads, crop areas and rivers to simulate the real environment.Obviously, robots can adapt to crop irrigation in different environments. 
 \begin{figure}[thpb]
	\centering
	\includegraphics[width=1\linewidth]{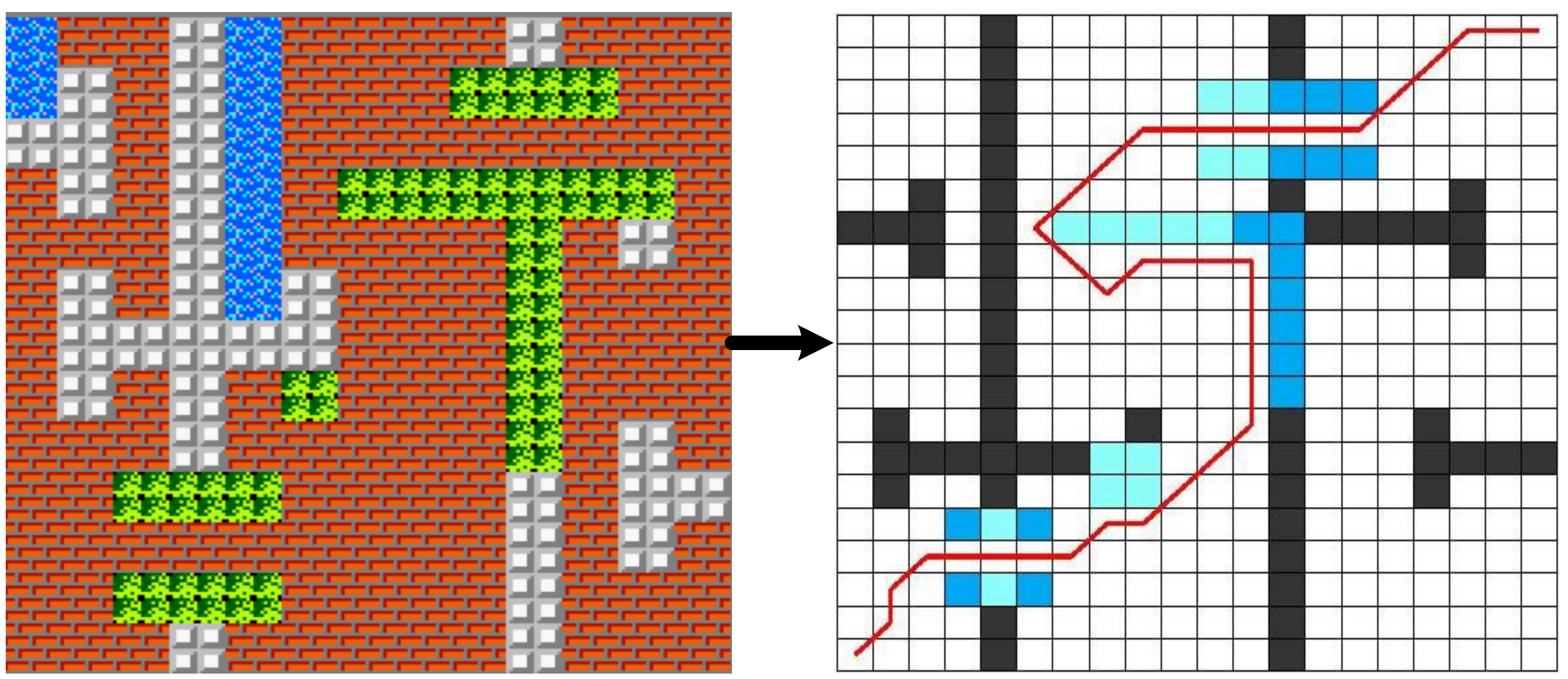}
	\caption{Flow chart of simulation experiment. We first transform the agricultural environment into a color simulation map. What's more, we use the algorithm to generate the grid map.}
	\label{fig:architecture}
\end{figure}

For each scenario, we randomly located the starting and target positions of the robot. Then, we ran the algorithm to find the shortest path between two specific locations. We can also determine whether the shortest path can be calculated by observing the curve trend of the shortest path length.

The performance of program in Matlab can be improved by modifying the parameters. For instance, the coefficients of related factors, generation of ants, and pheromone adjustment. We used the training scene to generate a applicable set of parameters. Next, we considered the adjustment process for these parameters.

We used the double maze scene to generate a proper number of iterations (adaptability to the scenario proposed by Shaw et al.). Furthermore, we changed the number $M\in \left\{ 100,150,200,250 \right\}$ and generation $K\in \left[ 0,100 \right]$ of ants to find the best global path. Obviously, high iterations tented to provide a better planning route, but it also increases the calculation amount and time. 

\begin{figure}[thpb]
	\centering
	\includegraphics[width=1\linewidth]{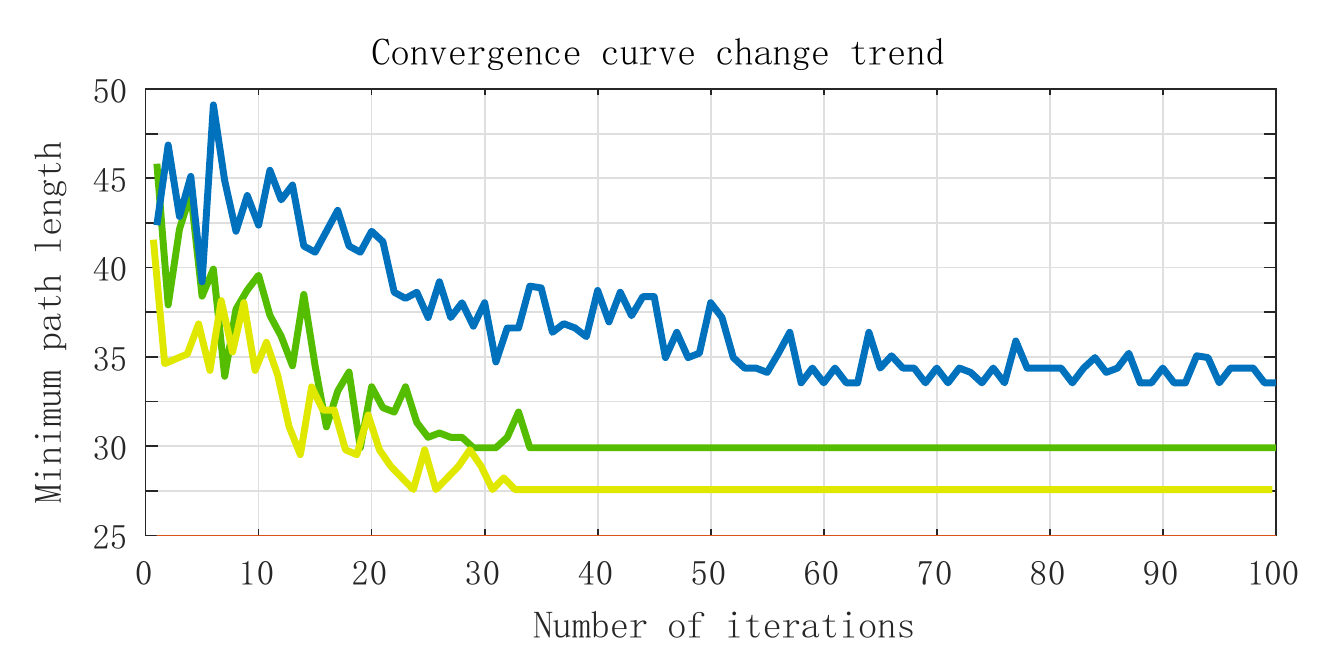}
	\caption{ Convergence curve of typical path obstacle graph. As the number of iterations increases, the generated path length change graph. The convergence curve is given. Select three kinds of path obstacle diagrams: corridor wave region, dense region and square spiral region to show. }
	\label{fig:architecture}
\end{figure}    

As presented in Fig. 4, the convergence curve of the most typical path barrier graphs shows a downward trend, albeit to some fluctuation. It is consistent with the previous statement that high iterations tented to provide a better planning route. Besides, each obstacle graph becomes stable when the number of iterations reaches about 100. So we set  $K=100$ to imporve the calculation efficiency of the algorithm. In the areas with more obstacles such as brick pattern and complex2, the operation results achieve stability after 30 iterations. It indicates that the improved algorithm is especially suitable for complex agricultural environment.

The pheromones can be adjusted by altering the values of $Alpha$, $Q$ and $Rho$ in the program. They represented the importance of pheromones, the enhancement coefficient and the evaporation coefficient respectively. The movement of ants will become increasingly accurate with the increase of $Alpha$. But when it rises to a certain value, the robot behaves abnormally. Especially in a small area, it tends to move back and forth repeatedly and become greatly affected by obstacles. This phenomenon can be seen in the maps[14]. In contrast, adjusting pheromones has little influence on results. It aims at improving the computational sensitivity. Generally, setting $Alpha=1$, $ Q=1$ and $Rho=2$ in the algorithm is suitable for most cases and achieves better performance.

\begin{figure*}[thpb]
	\centering
	\includegraphics[width=0.9\linewidth]{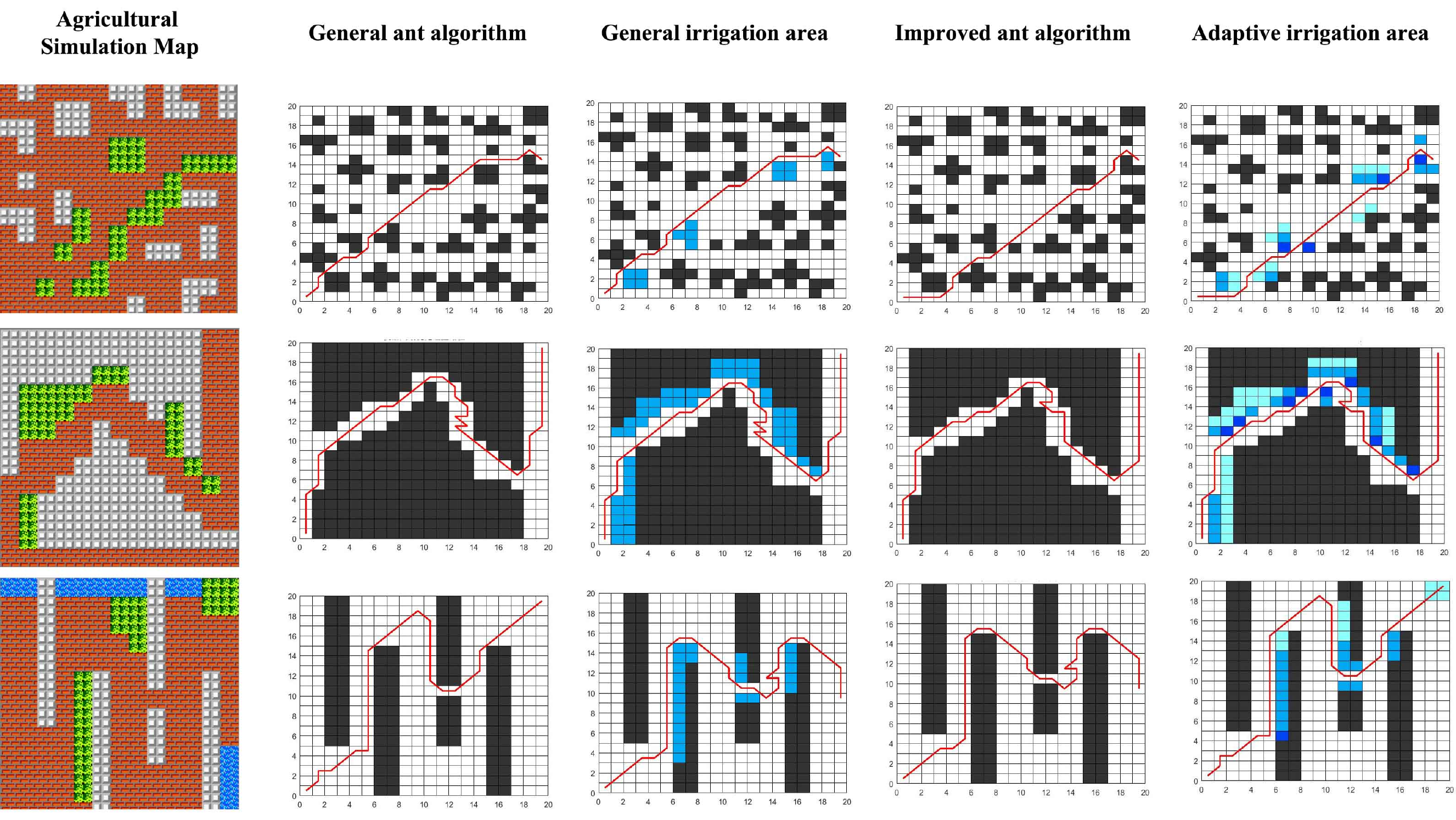}
	\caption{Performances comparation of general ant colony algorithm and improved algorithm in path planning and adaptive irrigation. The first column of the figure is the agricultural simulation map. The second and fourth columns of the image show the robot's irrigation path during the first and second round. The comparison of the third and fifth images indicates that the deeper the color is, the larger the irrigation volume is.}
	\label{fig:architecture}
\end{figure*}
 
As presented in Fig. 5, the first column is the generated agricultural simulation map. We compare the general path planning algorithm with the improved algorithm through comparative experiments. It can be seen from the figure that the improved algorithm can obtain more agricultural information and plan the adaptive irrigation area. For example, crops planted far away from the river need more water and are darker than those near the river. We simulated for five kinds of environment, namely (a) corridor waves, (b) dense area,  (e) front and rear[14]. Robots in all three environments are able to complete cruise and adaptive irrigation tasks.
\subsection{Field experiments}
The experimental base of the School of Tropical Agriculture and Forestry is an educational practice base. It lies near the southwest gate of Hainan University (In Fig. 6). Various kinds of agricultural crops and landscape plants are raised in this agricultural base. Therefore, the soil features the characteristics such as block structures and flake structures that are consistent with the actual agricultural environment. Besides, it also includes aggregate structure which is suitable for crop growth in the greenhouse. Different soil characteristics have different requirements for sprinkler intensity. We chose to conduct field experiments here to thoroughly test the validation of the robot and algorithm.\\
\begin{figure}[thpb]
	\centering
	\includegraphics[width=1\linewidth]{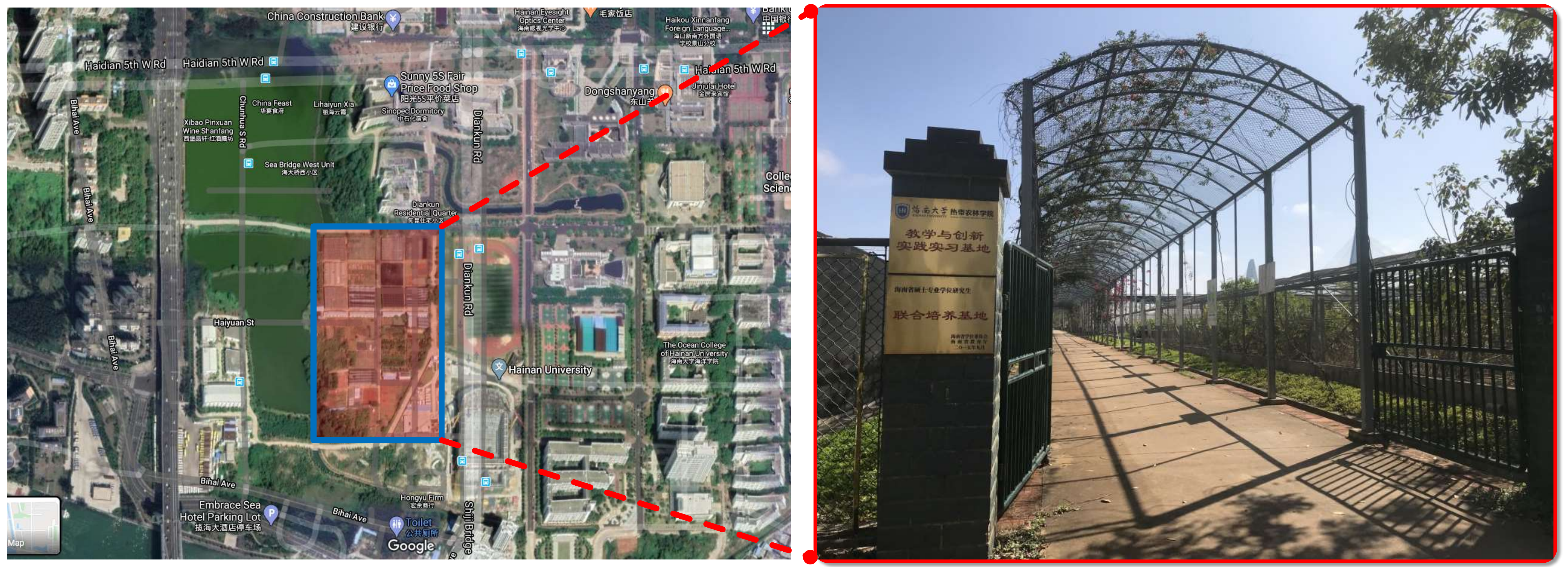}
	\caption{The experimental site: the experimental base of Tropical Agriculture and Forestry College of Hainan University}
	\label{fig:architecture}
\end{figure}

We developed an adaptive irrigation robot. As presented in Fig. 7, the tires, sprinkler device and navigation equipment were improved in accordance with the agricultural environment. Firstly, we used tires instead of crawler, which was easier to adjust the moving direction. Secondly, we designed horizontal and vertical patterns on the tire’s surface to increase the friction and prevent the splash of mud and water. In addition, we added shock absorbers at the bottom of the robot to make the movement smoother. 

\begin{figure}[thpb]
	\centering
	\includegraphics[width=1\linewidth]{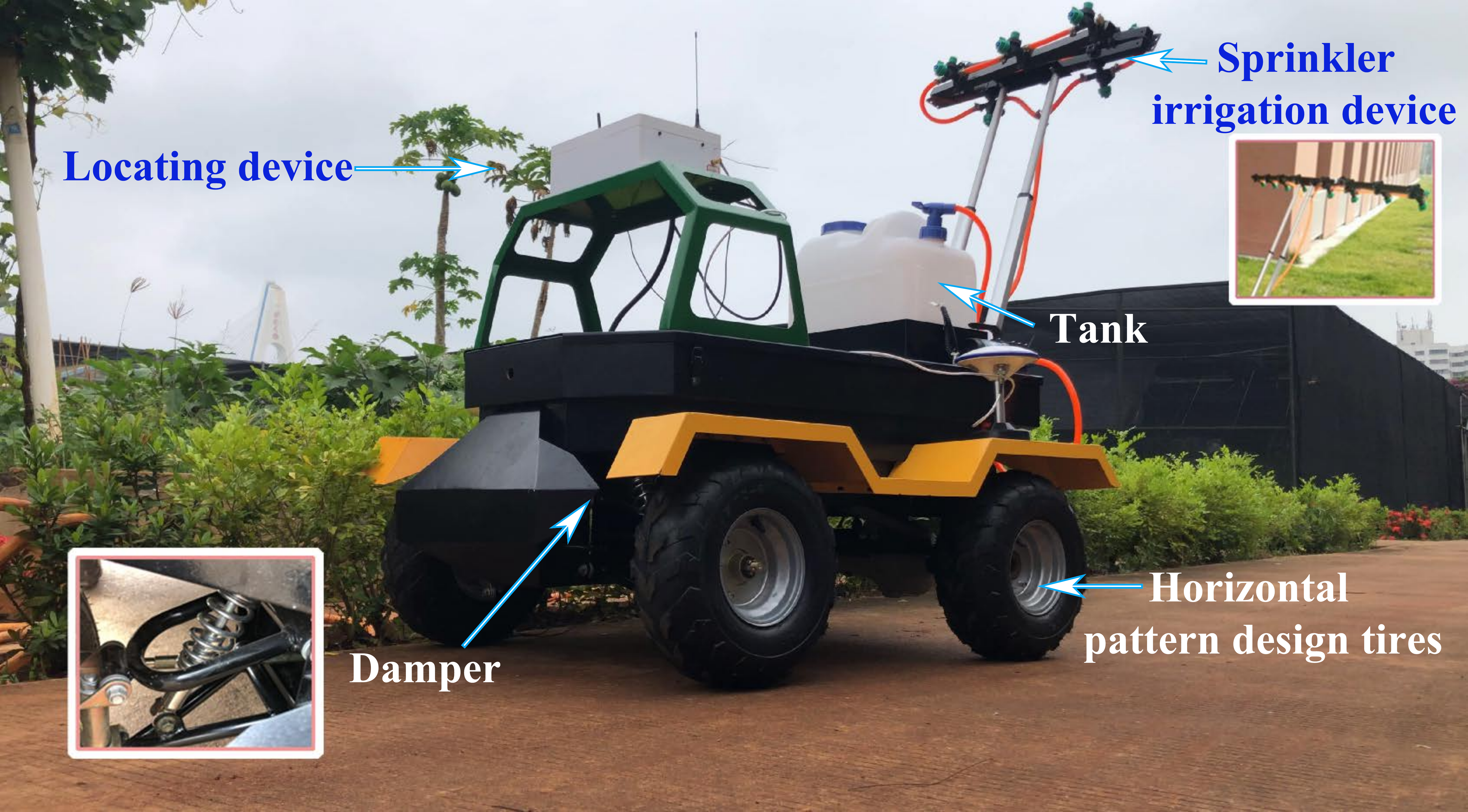}
	\caption{Adaptive irrigation robot.It is composed of foldable sprinkler device, shock absorber, horizontal and vertical pattern design tires,navigation system and water tank. }
	\label{fig:architecture}
\end{figure}
In the steering gear control system, the turning radius of the mobile robot is 0.5 m and the maximum forward / backward speed is 0.7 m/s. The experiments were performed in a real agricultural environment. We chose a rugged and narrow environment as an example. Mobile robots can perform tasks satisfactorily in their work-space.
\begin{figure}[thpb]
	\centering
	\includegraphics[width=1\linewidth]{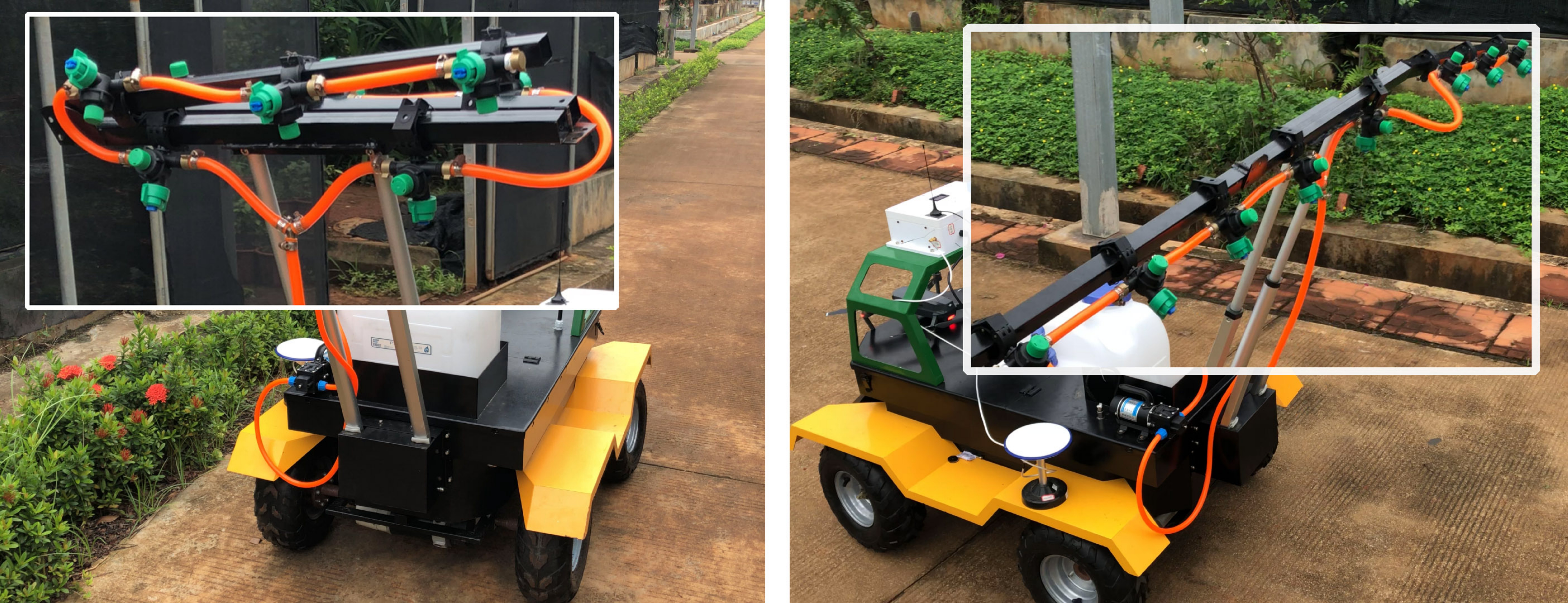}
	\caption{The sprinkler device. The left picture shows the shape of the sprinkler when folded, which is specific ally designed to avoid obstacles in narrow areas. On the right is an unfolded shape. It is able to expand both horizontally and vertically and increases the volume available for irrigation.}
	\label{fig:architecture}
\end{figure}

As presented in Fig. 8, the sprinkler device was foldable. It enabled the robot to pass through narrow areas and expanded the working area when unfolded. We installed 16 small-range sprayer on the device for irrigation. This type of sprinkler is not only suitable for small-scale irrigation,  but also completes the task dust cleaning. Regardless of the spray angle, the irrigation intensity is basically the same. This characteristic is extremely beneficial to ensure the spray uniformity of the system. The robot is equipped with a 20L water tank, which ensures sufficient water supply without adding too much weight.
\begin{figure}[thpb]
	\centering
	\includegraphics[width=1\linewidth]{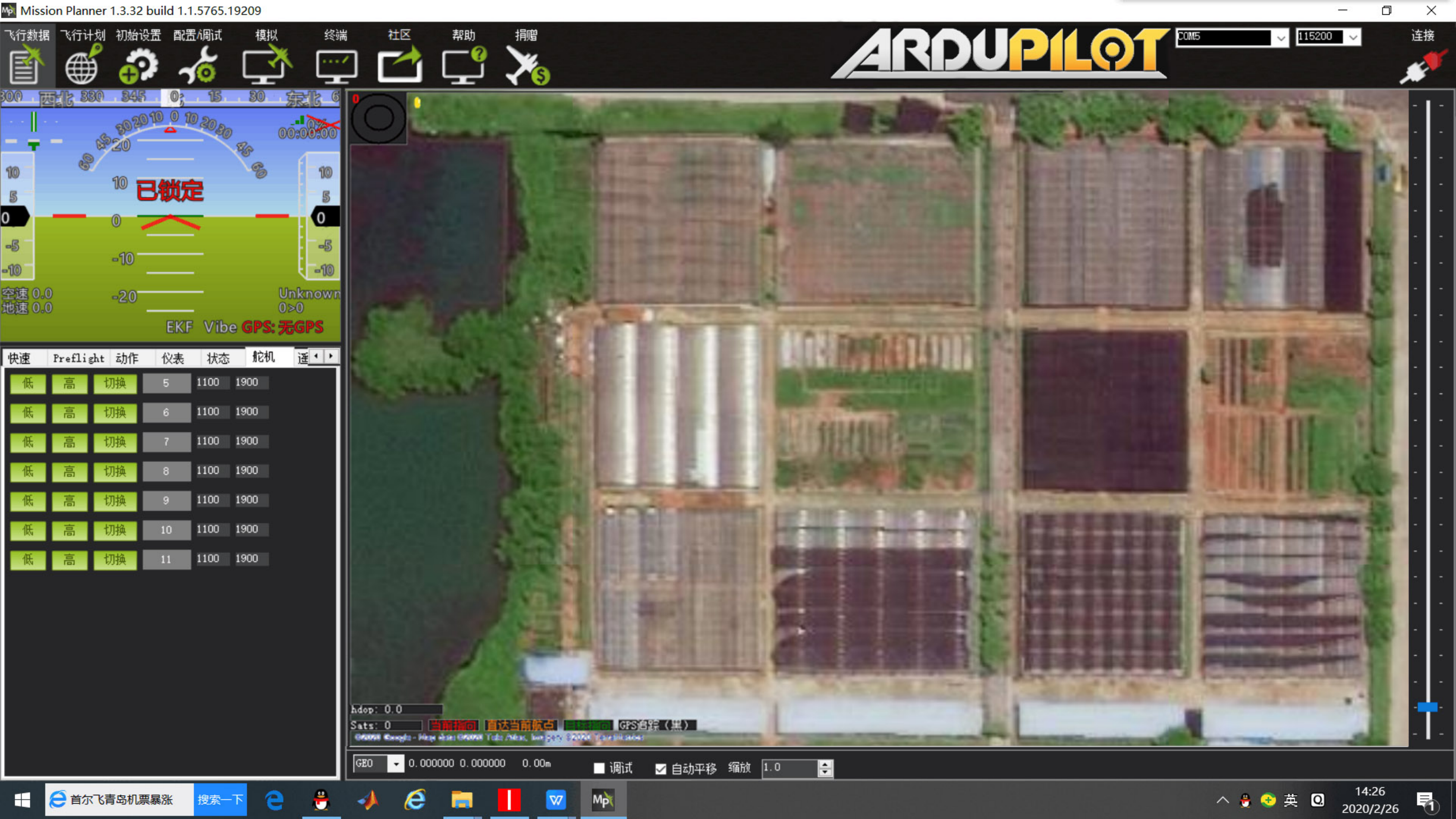}
	\caption{The simulated cruise map developed on Missionplanner}
	\label{fig:architecture}
\end{figure}

As presented in Fig. 9, on the left side of the navigation module is the steering gear control system, where cruise conditions such as the speed of the robot are shown. After connecting the robot to this system,  we were able to see the actual situation in front of the robot in the upper left corner of the interface. Besides, the navigation path and position of the robot will be displayed on the right interface. We can enlarge the location map on the right for precise positioning. The navigation system also supports manual changes of the robot's motion route.

As presented in Fig. 10, the first column is the actual environment map. After simulating the agricultural environment, an algorithm is used to generate a robot navigation path. The robot selects a path with more crops to carry out navigation and irrigation tasks. The comparative experiments are on the third and fourth columns. The results show that compared with the general ant colony algorithm, the improved algorithm enables the robot to collect more agricultural information without significantly increasing the working distance, which improves the effectiveness of the robot's navigation.

The improved algorithm takes the degree of drought into account. Therefore, the agricultural robot can carry out adaptive sprinkler irrigation according to the regional agricultural situation. Reasonable irrigation can not only save water, but also make crops have better growing environment.Similarly, if other factors need to be tested, repeating the previous steps to adjust the corresponding parameters, the improved algorithm is able to achieve the goal of precision irrigation.

We observe three important characteristics of the improved ant colony algorithm based on Bayesian theory:  

1) It is suitable for the rough and thorny agricultural environment. For example, narrow corridors, tunnel dogleg, tunnel twisted and tunnel. 

2) In the process of path selection, the improved ant algorithm does not necessarily choose the shorter path first, but opts for the shorter path under the premise that more information can be obtained. 

3) The algorithm has higher efficiency. It takes less time to generate paths in all planning problems by using benchmarks, to be precise, around 40 seconds. 

In conclusion, the ant colony algorithm based on Bayesian theory makes the robot well-performed in diverse kinds of environment. It also achieves the precision irrigation and ensures the high validity of each cruise.
\begin{figure*}[thpb]
	\centering
	\includegraphics[width=1\linewidth]{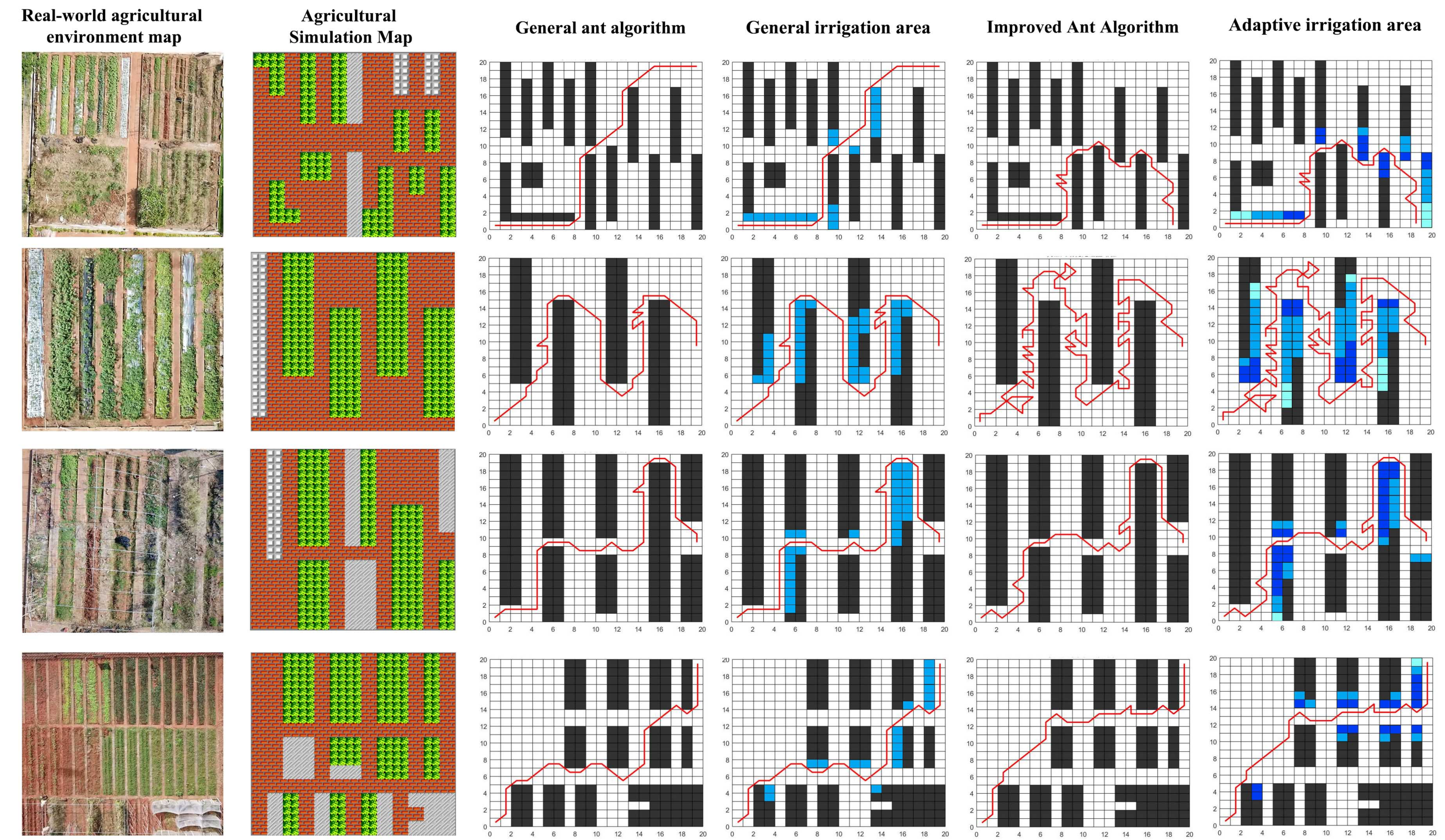}
	\caption{Comparison between classical ant colony algorithm and improved algorithm in the real agricultural environment: Compared with the original one, the path selected by the improved algorithm covered more areas without significantly increasing the distance.}
	\label{fig:architecture}
\end{figure*}
\section{CONCLUSION AND FUTURE WORK}

In this work, we designed an adaptive irrigation robot suitable for the agricultural environment, and then we improved the path planning algorithm. It had two main merits. First of all, the algorithm made it possible to carry out tasks based on actual agricultural conditions, which guaranteed the effectiveness of each round of cruise. We found out he robot performs better in the narrow and rugged road. Secondly, the agricultural robots and algorithms proposed in this paper have good variability. We are able to figure out the optimal probability value by adding mutation operators or directly analyzing the problem. As long as the corresponding influence coefficient is determined, other environmental factors such as light intensity, temperature, humidity can be preferentially investigated.  

Currently, we are able to simulate actual agricultural scenarios. The development of robot and navigation software has also been completed. However, there are still limitations in this work. The control between software and robots as well as the irrigation device has not been fully automated.

Our future work includes the following three aspects. Firstly, we are going to improve the automatic performance of the whole robotic system. Secondly, we will introduce convolutional neural network to path planning to realize automatically setting of coefficient. Finally, we will further expand the application of this method combined with cloud robotics[15, 16] and Artificial Intelligence. 

\addtolength{\textheight}{-12cm}


\begin{thebibliography}{99}
\bibitem{1} Miao Q and Shi H,`` Field Assessment of Basin Irrigation Performance in Hetao, Inner Mongolia", Irrigation \& Drainage Systems Engineering, Vol.6, No. 3 ,2017.
\bibitem{2}Xingye Zhu, Prince Chikangaise, Weidong Shi, Wen-Hua Chen, Shouqi Yuan, ``Review of intelligent sprinkler irrigation technologies for remoteautonomous system", Loughborough University Institutional Repository, Vol.11, No. 1, PP. 23-28, 2018.
\bibitem{3} Nor Adni Mat Leh, Muhammad Syazwan Ariffuddin Mohd Canadian, Zuraida Muhammad and Nur Atharah Kamarzaman,`` Smart Irrigation System Using Internet of Things", 2019 IEEE 9th International Conference on System Engineering and Technology (ICSET), Shah Alam, Malaysia, 2019, pp. 96-101.
\bibitem{4}Sajjad Yaghoubi, Negar Ali Akbarzadeh, Shadi Sadeghi Bazargani, Sama Sadeghi Bazargani, Marjan Bamizan, ``Autonomous Robots for Agricultural Tasks and Farm Assignment and Future Trends in Agro Robots", International Journal of Mechanical \& Mechatronics Engineering, Vol.13, No.3, 2013.
\bibitem{5}Francisco Rubio, Francisco Valero, Carlos Llopis-Albert, ``A review of mobile robots: Concepts, methods, theoretical framework, and applications", International Journal of Advanced Robotic Systems, Vol. 16, No. 2, 2019.
\bibitem{6} T. W. Bank, ``Globally, 70\% of Freshwater is Used for Agriculture", 2017. [Online]. Available:  https://blogs.worldbank.org/opendata/chart-globally-70-freshwater-used-agriculture
\bibitem{7}Qingfeng Miao, Haibin Shi, José M. Gonçalves and Luis S. Pereira,`` Basin Irrigation Design with Multi-Criteria Analysis Focusing on Water Saving and Economic Returns: Application to Wheat in Hetao, Yellow River Basin”, application to wheat in Hetao, Yellow River basin. Water, Vol. 10, No. 1, pp. 68, 2018. 
\bibitem{8}Angelopoulos, Constantinos Marios, Sotiris Nikoletseas, and Georgios Constantinos Theofanopoulos,`` A Smart System for Garden Watering using Wireless Sensor Networks”, MobiWac 11, Proceedings of the 9th ACM International Symposium on Mobility Management and Wireless Access, New York, NY, USA, 2011, pp. 167-170.
\bibitem{9}O.P.Bound, U.C. Adie, O.M. Ikumapayi, J.O. Akinyoola, A.A. Aderoba,`` Architectural design and performance evaluation of a ZigBee technology based adaptive sprinkler irrigation robot” Contents lists available at ScienceDirect,Vol. 160, pp. 168-178, 2019
\bibitem{10}Adeodu A. O, Bodunde O. P, Daniyan I. A, Omitola O. O, Akinyoola J. O, Adie U. C” Development of an autonomous mobile plant irrigation robot for semi structured environment”, Procedia Manufacturing, Vol. 35,  pp. 9- 15, 2019.
\bibitem{11}Guoyu Zuo, Peng Zhang, Junfei Qiao,`` Path Planning Algorithm Based on Sub-Region for Agricultural Robot", International Asia Conference on Informatics, vol. 2, pp. 179-200, 2010.
\bibitem{12}I.A.Hameed, A.la Cour-Harbo, O.L.Osen,  ``side-to-side 3D coverage path planning approach for agricultural robots to minimize skip/overlap areas between swaths",  Robotics and Autonomous Systems, Vol. 76, pp. 36-45, 2016.
\bibitem{13}Mogens Graf Plessen, ``Coupling of crop assignment and vehicle routing for harvest planning in agriculture", Artificial Intelligence in Agriculture, vol. 2, pp. 99-109, 2019.
\bibitem{14}Marco A. Contreras-Cruz, Victor Ayala-Ramirez, Uriel H. Hernandez-Belmonte, ``Mobile robot path planning using artificial bee colony and evolutionary programming", Applied Soft Computing, vol. 30, pp. 319-328, 2015.
\bibitem{15}B. Liu, L. Wang, M. Liu, and C. Xu, ``Lifelong federated reinforcement learning: a learning architecture for navigation in cloud robotic systems", IEEE Robotics and Automation Letters (RA-L), vol. 4, No. 4, pp. 4555 - 4562, 2019.
\bibitem{16}B. Liu, L. Wang, M. Liu, and C. Xu, ``Federated Imitation Learning: A Novel Framework for Cloud Robotic Systems with Heterogeneous Sensor Data", arXiv preprint, arXiv:1911.02685, 2019.
\end{thebibliography}
\end{document}